\documentclass[final]{cvpr}

\usepackage{times}
\usepackage{epsfig}
\usepackage{graphicx}
\usepackage{amsmath}
\usepackage{amssymb}
\usepackage{bm}
\usepackage{fixltx2e}
\usepackage{placeins}
\usepackage{gensymb}
\usepackage{tablefootnote}
\usepackage{xcolor}
\usepackage[multiple]{footmisc}
\usepackage[font=footnotesize]{caption}
\usepackage[pagebackref=true,breaklinks=true,letterpaper=true,colorlinks,bookmarks=false]{hyperref}
\usepackage[colorlinks]{hyperref}

\begin{document}

\title{LEGAN: Disentangled Manipulation of Directional Lighting and Facial Expressions whilst Leveraging Human Perceptual Judgements \thanks{*Work done while at Affectiva}}
\author{\parbox{16cm}{\centering
    {\large Sandipan Banerjee$^1$, Ajjen Joshi$^1$, Prashant Mahajan*$^2$, Sneha Bhattacharya*$^3$, Survi Kyal$^1$ and Taniya Mishra*$^4$}\\
    {\normalsize
    $^1$Affectiva, USA, $^2$Amazon, USA, $^3$Silver Spoon Animation, USA, $^4$SureStart, USA\\
        \tt\small \{firstname.lastname\}@affectiva.com, \tt\small prmhp@amazon.com, \tt\small snehabhattac@umass.edu, \tt\small taniya.mishra@mysurestart.com
    }}
}

\renewcommand\footnotemark{}
\maketitle

\begin{abstract}
\vspace{-0.5cm}
Building facial analysis systems that generalize to extreme variations in lighting and facial expressions is a challenging problem that can potentially be alleviated using natural-looking synthetic data. Towards that, we propose LEGAN, a novel synthesis framework that leverages perceptual quality judgments for jointly manipulating lighting and expressions in face images, without requiring paired training data. LEGAN disentangles the lighting and expression subspaces and performs transformations in the feature space before upscaling to the desired output image. The fidelity of the synthetic image is further refined by integrating a perceptual quality estimation model, trained with face images rendered using multiple synthesis methods and their crowd-sourced naturalness ratings, into the LEGAN framework as an auxiliary discriminator. Using objective metrics like FID and LPIPS, LEGAN is shown to generate higher quality face images when compared with popular GAN models like StarGAN and StarGAN-v2 for lighting and expression synthesis. We also conduct a perceptual study using images synthesized by LEGAN and other GAN models and show the correlation between our quality estimation and visual fidelity. Finally, we demonstrate the effectiveness of LEGAN as training data augmenter for expression recognition and face verification tasks.
\end{abstract}

\vspace{-0.5cm}
\section{Introduction}
\label{sec:intro}
\vspace{-0.2cm}

Deep learning \cite{DLNature} has engendered tremendous progress in automated facial analysis, with applications ranging from face verification \cite{VGG,VGGFace2,DosDonts} to expression classification \cite{FECNet}. However, building robust and accurate models that generalize effectively in-the-wild is still an open problem. A major part of it stems from training datasets failing to represent the ``true'' distribution of real world data \cite{MegaFace,MF2,UG2,StrikewithPose} (\eg extreme lighting conditions \cite{BeveridgeLighting,GBU,KlemenEffect}); or the training set may be non-uniformly distributed across classes, leading to the long-tail problem \cite{MasiAug}.

One way to mitigate the imbalance problem, shown to work in multiple domains \cite{masiFG17,SREFI2,SBeery,XRayGAN}, is to introduce synthetic samples into the training set. Many approaches for generating synthetic data exist \cite{Bitouk08,SREFI1,HassFront}, none as successful as GANs \cite{GAN} in generating realistic face images \cite{BEGAN,ProgressiveGAN,StyleGen,StarGAN}. Thus, in this work, we design a GAN model for synthesizing variations of an existing face image with the desired illumination and facial expression, while keeping the subject identity and other attributes constant. These synthetic images, when used as supplemental training data, can help build facial analysis systems that better generalize across variations in illumination and expressions.

\begin{figure}[t]
\centering
   \includegraphics[width=1.0\linewidth]{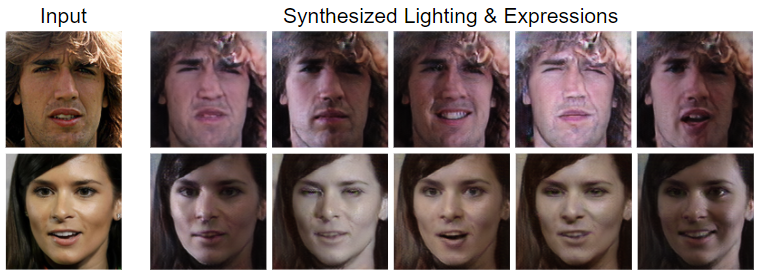}
   \caption{LEGAN jointly manipulates lighting and expression in face images while preserving the subject identity of the input.}
\label{fig:Teaser}
\vspace{-0.6cm}
\end{figure}

\begin{figure*}[t]
\centering
   \includegraphics[width=0.9\linewidth]{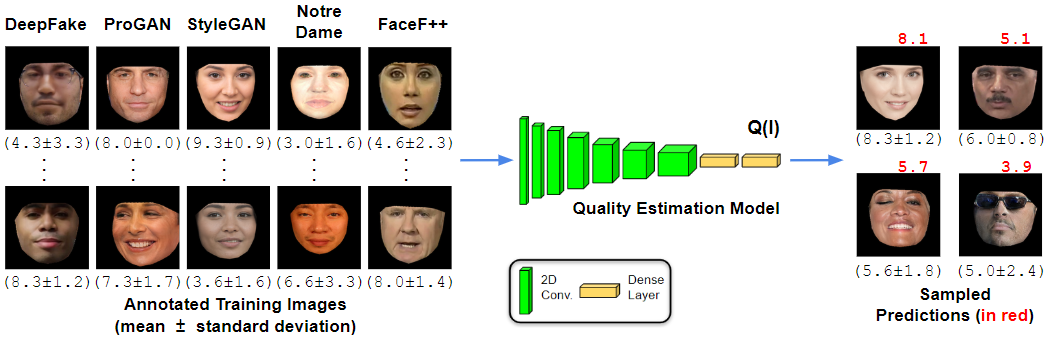}
   \caption{Our quality estimation model is trained on synthetic face images, varying in gender, ethnicity, lighting and facial attributes, generated with five methods \cite{DeepFakeChallenge,ProgressiveGAN,StyleGen,SREFI2,FaceFPP} and their crowd-sourced perceptual ratings. To account for the subjective nature of human perception, our model utilizes a margin based regression loss to learn representations. This trained model can then generate naturalness predictions (Q(I)) of unseen face images.}
\label{fig:Qual_Model}
\vspace{-0.5cm}
\end{figure*}

One drawback of GAN-based face generation is the absence of an accurate and automated metric to judge the perceptual quality of synthesized images \cite{HumanJudgement,BorjiGANEval}. In order to solve this problem, we also introduce a quality estimation model that can serve as a cheap but efficient proxy for human judgment while evaluating naturalness of synthetic face images. Instead of generating a single score for a distribution of synthetic images \cite{Hype, FID, IS} or for image pairs \cite{ZhangCVPR18,PieAPP}, our goal is to infer an image-quality score on a continuum for a single synthetic face image. With this in mind, we run an Amazon Mechanical Turk (AMT) experiment where turkers are instructed to score the naturalness of synthetic face images, generated using different 3D-model \cite{SREFI2} and GAN-based \cite{ProgressiveGAN,StyleGen,FaceFPP,DeepFake} synthesis approaches. We then build a feed forward CNN to learn representations from these images that map to their corresponding perceptual rating, using a margin based regression loss.

In addition to a traditional discriminator \cite{GAN}, this trained quality model is then used as an auxiliary discriminator in the synthesis framework, named LEGAN (Lighting-Expression GAN), that we propose in this paper. Instead of intertwining the two tasks \cite{StarGAN}, LEGAN decomposes the lighting and expression sub-spaces using a pair of hourglass networks (encoder-decoder) that generate transformation masks capturing the intensity changes required for target generation. The desired output image is then synthesized by a third hourglass network from these two masks.

To demonstrate the effectiveness of LEGAN, we qualitatively and quantitatively compare its synthesized images with those produced by two popular GAN based models \cite{StarGAN, StarGAN2} using objective metrics like FID \cite{FID}, LPIPS \cite{ZhangCVPR18}, SSIM \cite{SSIM} and face match score. We also conduct a human rater study to evaluate the perceptual quality of LEGAN's images and the contribution of the quality based auxiliary discriminator towards hallucinating perceptually superior images. Finally, we show the efficacy of LEGAN as training data augmenter by improving the generalizability of face verification and expression recognition models.

\vspace{-0.1cm}
\section{Related Work}
\label{sec:related}
\vspace{-0.1cm}
{\bf Face Synthesis}: Early approaches \cite{Bitouk08,ACCV14} focused on stitching together similar looking facial patches from a gallery to synthesize a new face. Manipulating the facial shape using 3D models \cite{Bulat3D,USC3DMM,MasiAug,SREFI2} or deep features \cite{Belanger,DFI} is another popular approach to generate new views. In recent times however researchers have pre-dominantly focused on using GANs \cite{GAN} for synthesis, where an upsampling generator hallucinates faces from a noise vector, either randomly sampled from a distribution \cite{GAN,DCGAN,ProgressiveGAN,StyleGen} or interpreted from a different domain \cite{Speech2Face}. An existing face can also be encoded and then upsampled to obtain the desired attributes \cite{pix2pix,TPGAN,Voice2Face,SREFI3,FSGAN,VishalSketch,Gecer_2019_CVPR,CycleGAN,PairedCycleGAN,StarGAN}.

\begin{figure*}[t]
\centering
   \includegraphics[width=0.9\linewidth]{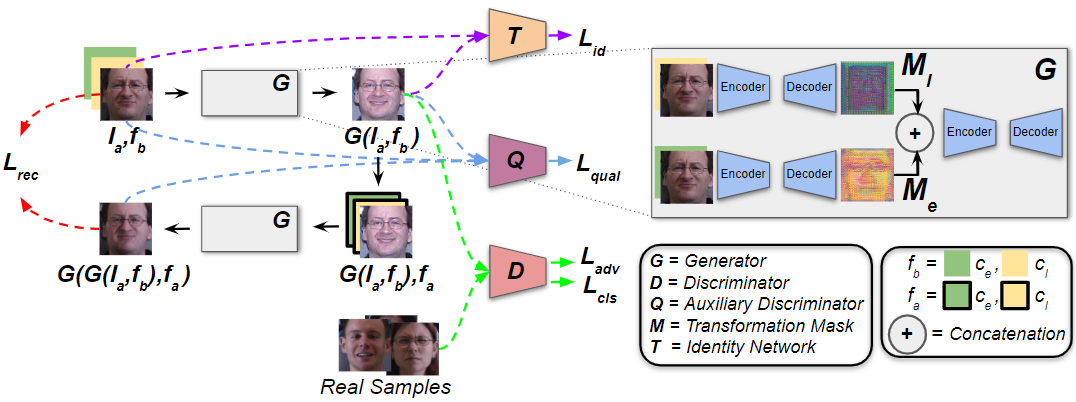}
   \caption{During training, $G$ takes an image $I_{a}$ and target expression and lighting tensor $f_{b}$ as input and disentangles the feature sub-spaces using a pair of hourglass networks generating transformation masks $M_{e}$ and $M_{l}$ that are concatenated and passed through a third hourglass to hallucinate the target output $G(I_{a},f_{b})$. To eliminate the need of paired training data, we augment $G(I_{a},f_{b})$ with the source expression and lighting tensor $f_{a}$ and pass it through the same generator to reconstruct the input $G(G(I_{a},f_{b}),f_{a})$, to compute the reconstruction error $L_{rec}$. Moreover, with the help of $D$ and the quality based auxiliary discriminator $Q$, we calculate the adversarial ($L_{adv}$), feature classification ($L_{cls}$) and quality ($L_{qual}$) losses respectively. Utilizing the identity network $T$, we compute the identity loss $L_{id}$. During testing, only $G$ is required to generate synthetic images.}
\label{fig:LEGAN_Model}
\vspace{-0.5cm}
\end{figure*}

\noindent {\bf Editing Expressions}: Research in this domain started with modeling skin-muscle movements \cite{Pictorial} for different facial expressions or swapping facial patches based on visual proximity \cite{Bitouk08}. With the advent of 3D face models, researchers used static \cite{MasiAug} or morphable models \cite{blanz2003face,Perception} to manipulate facial expressions with a higher degree of realism. Recently, the use of VAEs \cite{GangHua} and adversarial image-to-image translation networks have become extremely popular for editing facial expressions \cite{pix2pix,CycleGAN,StarGAN}, with or without paired data for training. Some of these models use attention masks \cite{GANimation}, facial shape information \cite{3DGuided,HighFidelity} or exemplar videos \cite{Siarohin_2019_NeurIPS} to guide the model in this task.

\noindent {\bf Editing Lighting}: While methods like histogram equalization \cite{AHE,CLAHE} and gamma correction can shift the global luminance distribution and color encoding of an image, they cannot manipulate the direction of the light source itself. An early method \cite{L_PAMI09} utilizes spherical harmonics to manipulate the directional lighting in 3D. In \cite{eccv10}, local linear adjustments are performed on overlapping windows to change the lighting profile of an image. Deep learning based approaches have also been proposed where the reflectance, normal and lighting channels are disentangled and edited to relight images \cite{SfSNet,NeuralEdit,L_Mass,Neuro20}. Alternatively, the desired lighting can be passed to an encoder-decoder pair as the target for lighting manipulation in the input image \cite{DeepPortrait,PortraitRelit19,DeepRef,nestmeyercvpr20}. Recently, joint facial pose, lighting and expression manipulation has been proposed in \cite{DiscoFaceGAN,CONFIG} where an input image can be manipulated by changing its attributes in feature space leveraging 3D parameters or latent information from synthetic images during training.

\noindent {\bf Quality Estimation of Synthetic Face Images}: Synthetic image quality is commonly evaluated using metrics like the Inception Score \cite{IS} or FID \cite{FID}, which compare statistics of real and synthetic feature distributions, and output a single score for the whole distribution rather than the individual image. The features themselves are extracted from the Inception-v3 model \cite{Inceptionv3}, usually pre-trained on objects from \cite{ImgNet}, and not specifically faces. As these metrics do not take into account human judgements, they do not correlate well with perceptual realism \cite{InceptionNote,BorjiGANEval}. Consequently, researchers run perceptual studies to score the naturalness of synthetic images \cite{ColorZhang,SREFI1}. These ratings are also used to design models that measure distortion between real and synthetic pairs \cite{ZhangCVPR18,PieAPP} or the coarse realism (`real' vs `fake') of a synthetic image \cite{Hype}. None of these evaluation models however are designed specifically for face images. Recently, \cite{HumanJudgement} proposed a metric to rate the perceptual quality of a single image by using binary ratings from \cite{Hype} as ground truth for synthetic face images generated by \cite{ProgressiveGAN,StyleGen}. However, their regression based model is trained on only 4,270 images and thus insufficient to reliably model the subjective nature of human judgements.

Unlike these methods, we build a synthetic face quality estimation model by leveraging perceptual ratings of over 37,000 images generated using five different synthesis techniques \cite{ProgressiveGAN,StyleGen,FaceFPP,DeepFakeChallenge,SREFI2}. Our quality model takes into account the variability in human judgements and generate a realism score for individual images rather than the whole set. We leverage this model as an auxiliary discriminator in the LEGAN framework for simultaneous lighting and facial expression manipulation. This novelty together with LEGAN's feature disentanglement improves the naturalness of the hallucinated images. Additionally, we do not require external 3DMM information or latent vectors during training nor do we need to fine-tune our model during testing on input images \cite{DiscoFaceGAN,CONFIG}.

\vspace{-0.3cm}
\section{Quality Estimation Model}
\label{sec:quality}
\vspace{-0.1cm}
Our quality estimator model is trained with synthetic face images assembled and annotated in two sequential stages, as described below.

{\bf Stage I}: We first generate 16,507 synthetic face images using the StyleGAN \cite{StyleGen} generator. These images are then annotated by labelers using Amazon Mechanical Turk (AMT) on a scale of 0 - 10 for naturalness, where a 0 rating represents an unnatural image and 10 a hyper-realistic one. The images are then binned into two broad groups - `unnatural' for AMT ratings between 0 - 5 and `natural' for 5 - 10. We extract descriptors for each image from the `avg\_pool' layer of the ResNet50 \cite{ResNet} model, pre-trained on VGGFace2 \cite{VGGFace2} and train a linear SVM \cite{SVM} with the extracted features of around 12,000 images from this dataset and use the rest for parameter tuning. Post training, we use this SVM as a rough estimator of naturalness.

{\bf Stage II}: In this stage, we perform the same AMT experiment again with a larger set of synthetic face images, collected from the following datasets:\\
1. {\bf FaceForensics++}\cite{FaceFPP} - we randomly sample 1000 frames from this dataset consisting of 1000 video sequences that have been manipulated with four automated face manipulation methods.\\
2. {\bf DeepFake}\cite{DeepFake,FaceSwap} - we use sampled frames from 620 manipulated videos of 43 actors from \cite{DeepFakeChallenge}.\\
3. {\bf ProGAN} \cite{ProgressiveGAN} - we generate 10,000 synthetic face images of non-existent subjects by training NVIDIA's progressively growing GAN model on the CelebA-HQ dataset \cite{ProgressiveGAN}.\\
4. {\bf StyleGAN} \cite{StyleGen} - we extract 100,000 hyper-realistic face images of non-existent subjects generated using the StyleGAN model that were pre-filtered for quality \cite{FilteredStyleGAN}.\\
5. {\bf Notre Dame Synthetic Face Dataset} \cite{SREFI2} - we randomly sample 163,000 face images, from the available 2M, of synthetic subjects generated using `best-fitting' 3D models.

To focus on near-frontal faces, we remove images with yaw over 15$\degree$ in either direction, estimated using \cite{Keke}. Since gender information is absent in most of the above datasets, we group the synthetic images using gender predictions from a pre-trained model \cite{HassGen}. Our trained SVM (from Stage I) is also used to rate the coarse naturalness of the collected images, using their ResNet50 features. We ensure balance in our synthetic dataset by sampling evenly from the natural and unnatural sets, as estimated by the SVM, and the perceived gender classes. To focus solely on the facial region, the pixels outside the convex hull formed by the facial landmarks, estimated using \cite{BulatLandmark}, of an image are masked. After the gender, facial yaw and naturalness based filtering, and the pre-processing step, we end up with 37,267 synthetic face images\footnote{Available here: \url{https://github.com/Affectiva/LEGAN_Perceptual_Dataset}} for our second AMT experiment.

Again, we ask Turkers to rate each image for naturalness on a scale of 0 - 10. Each image is shown to a Turker for 60 seconds to allow them time to make proper judgement even with slow network connection. We divide the full set of images into 72 batches such that each batch gets separately rated by 3 different Turkers. Post crowd-sourcing, we compute the mean ($\mu$) and standard deviation ($\sigma$) from the 3 scores and assign them as naturalness label for an image.

To train the quality estimation model, we use 80\% of this annotated data and the rest for validation and testing. For augmentation, we only mirror the images as other techniques like translation, rotation and scaling drastically change their appearance compared to what the Turkers examined. Our model downsamples an input image using a set of strided convolution layers with Leaky ReLU \cite{LeakyRelu} activation followed by two fully connected layers with linear activation and outputs a single realness scoring. Since both $\mu$ and $\sigma$ are passed as image labels, we try to capture the inconsistency in the AMT ratings (i.e. the subjective nature of human perception) by formulating a margin based loss for training. The model weights are tuned such that its prediction is within an acceptable margin, set to $\sigma$, from the mean rating $\mu$ of the image. The loss $L_{N}$ can be represented as:
\vspace{-0.2cm}
\begin{equation}
L_{N} = \frac{1}{n}\sum_{i=1}^{n}\left \|  \sigma - \left \|  \mu - Q(I_{i}) \right \|_{2}^{2} \right \|_{2}^{2}
\label{eq:L_Nat}
\end{equation}
where $n$ is the batch size, $Q(I_{i})$ is the model prediction for the $i$-th image in the batch. Since the model is trained on the mean rating $\mu$ (regression) as the target rather than fixed classes (classification), $L_{N}$ pushes the model predictions towards the confidence margin $\sigma$ from $\mu$.

\begin{figure}[t]
\centering
  \includegraphics[width=0.9\linewidth]{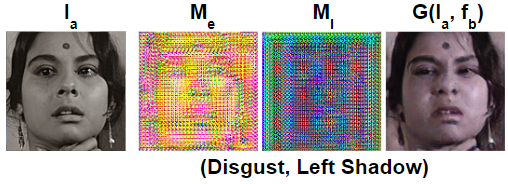}
  \caption{Sample LEGAN result $G(I_{a},f_{b})$, with generated expression $M_{e}$ and lighting $M_{l}$ masks for input $I_{a}$. The salient pixels for the translation task automatically `heat up' in $M_{e}$ and $M_{l}$, similar to flow maps.}
\label{fig:Gallery}
\vspace{-0.6cm}
\end{figure}

\begin{figure*}[t]
\centering
   \includegraphics[width=0.8\linewidth]{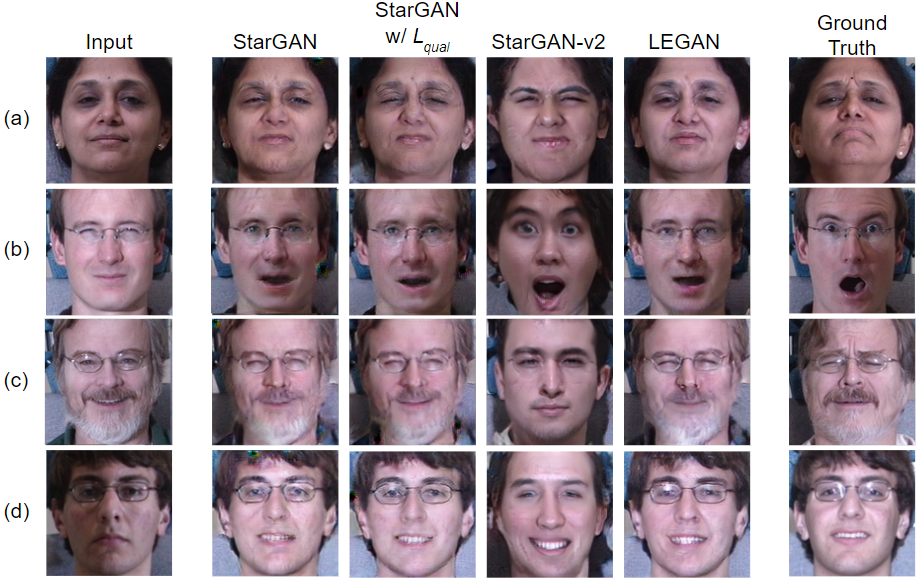}
   \caption{Sample results on MultiPIE \cite{MultiPIE} test images comparing LEGAN with popular GAN models. The target conditions are (a) (Disgust, Slight Left Shadow), (b) (Surprise, Slight Right Shadow), (c) (Squint, Ambient lighting), and (d) (Smile, Bright lighting). LEGAN hallucinates subtle muscle movements like nose wrinkles more prominently (top row) while preserving the subject identity. Additionally, our quality loss eliminates blob artifacts \cite{StyleGAN2} from synthesized images (bottom row, StarGAN vs. StarGAN w/ $L_{qual}$).}
\label{fig:Qual_Comp}
\vspace{-0.5cm}
\end{figure*}

\section{LEGAN}
\label{sec:legan}
We build LEGAN as a lightweight network that works with unpaired data, similar to the StarGAN family, to focus more on assessing the effect of our quality estimation model ($Q$) as a perceptual loss. Unlike \cite{DiscoFaceGAN,CONFIG}, LEGAN does not require additional networks to regress 3DMM parameters or fine-tuning during inference. We describe the architecture and objective functions of LEGAN in this section, an overview of which can be seen in Figure \ref{fig:LEGAN_Model}.
\vspace{-0.1cm}
\subsection{Architecture}
\vspace{-0.1cm}
{\bf Generator}: Our generator $G$, composed of three hourglass networks (encoder-decoder), starts with an input RGB face image $I_a$ and a target attributes vector $f_{b}$ that corresponds to expression and lighting conditions $c_{e}$ and $c_{l}$ respectively. The first hourglass receives $I_{a}$ concatenated with $c_{e}$ while the second one receives $I_{a}$ concatenated with $c_{l}$, thus disentangling the transformation task. Inside each hourglass, the concatenated tensor is downsampled using strided convolutions and then passed through a set of residual blocks \cite{ResNet} before being upsampled using pixel shuffling layers \cite{pixshuff}. Each convolution layer is followed by instance normalization \cite{InstanceNorm} and ReLU activation \cite{Relu} for non-linearity. These upsampled images are the transformation masks $M_{e}$ and $M_{l}$ that map the changes in pixel intensity required to translate $I_{a}$ to conditions specified in $f_{b}$. $M_{e}$ and $M_{l}$ are concatenated together and fed to the third hourglass to generate the output image $G(I_a,f_b)$. The objective of dividing the generation process into two stages and hallucinating the transformation masks is two fold - (a) easing the task of each hourglass by simply making it focus on registering the required expression or lighting changes instead of both registration and hallucination, and (b) making the transformation process more explainable, with salient pixels prominent in $M_{e}$ and $M_{l}$, as can be seen in Figure \ref{fig:Gallery}.

{\bf Discriminator}: The discriminator $D$ takes the output image $G(I_a,f_b)$ and predicts not only its realness score but also classifies its attributes $f_{b}$. $D$ is composed of strided convolution layers with Leaky ReLU \cite{LeakyRelu} activation that downsample the image to extract its encoded feature map. We use a patch discriminator \cite{pix2pix} that takes this encoded feature map and passes it through a single channel convolution to get the realness map $D_{src}$. This feature map is also operated by a conv layer with $k$ filters to get the attributes prediction map $D_{cls}$, where $k$ = no. of channels in $f_{b}$.

{\bf Auxiliary Discriminator}: We integrate the perceptual quality model $Q$, described in Section \ref{sec:quality}, into the LEGAN model graph to further refine the naturalness of the images synthesized by $G$. Unlike $D$, we do not train $Q$ jointly with $G$ but use the weights of a pre-trained snapshot.

{\bf Identity Network}: We also add a pre-trained identity preserving network \emph{T} to estimate the deviation of the output identity from that of the input. \emph{T} is trained offline on face images with different pose, expression and lighting and its weights are kept frozen throughout LEGAN's training.

\subsection{Loss Function}
{\bf 1. Adversarial Loss}: $D$ is trained to distinguish a real face image $I_{a}$ from its synthetic counterpart and judge the realness of the hallucinated image $G(I_a,f_b)$. To stabilize the gradients and improve quality, we use the WGAN \cite{wgan} based objective $L_{adv}$ for this task with a gradient penalty \cite{wgangp}, set as:
\vspace{-0.1cm}
\begin{multline}
L_{adv} = \mathbb{E}_{I_a}[D_{src}(I_a)] - \mathbb{E}_{I_a,f_b}[D_{src}(G(I_a,f_b))] - \\ \lambda_{gp}\mathbb{E}_{\hat{I}}[(\left \| \nabla_{\hat{I}}D_{src}(\hat{I}) \right \|_2 - 1)^2]
\label{eq:L_adv}
\end{multline}
where $\hat{I}$ is sampled uniformly from real and synthetic images and $\lambda_{gp}$ is an tunable parameter. While $D$ tries to minimize this to separate the synthetic from the real, $G$ tries to maximize it by fooling $D$.

{\bf 2. Classification Loss}: To ensure the target lighting and expression are correctly rendered by $G$ and enable LEGAN to do many-to-many translations, we formulate a classification loss using $D$'s predictions, in the form of $D_{cls}$. The loss $L_{cls}$ is computed as:
\vspace{-0.05cm}
\begin{multline}
L_{cls} = \mathbb{E}_{I_a,f_a}[-\log D_{cls}(f_a \mid I_a)] + \\ \mathbb{E}_{I_a,f_b}[-\log D_{cls}(f_b \mid G(I_a,f_b))]
\label{eq:L_cls}
\end{multline}
where $f_a$ and $f_b$ are the original and target attributes of an input image $I_a$.

{\bf 3. Identity Loss}: To preserve the subject identity without using paired data, we add an identity loss between the input and the translated output $G(I_a,f_b)$ by utilizing representations from \emph{T}. Both $I_a$ and $G(I_a, f_b)$ are passed through \emph{T} for feature extraction and we set the objective $L_{id}$ to minimize the cosine distance between these two features as:
\vspace{-0.1cm}
\begin{equation}
L_{id} = \mathbb{E}_{I_a,f_b}[1 - \frac{T(I_a) \cdot T(G(I_a,f_b))}{\left \| T(I_a) \right \|_{2}\left \| T(G(I_a,f_b)) \right \|_{2}}]
\label{eq:L_id}
\end{equation}

\noindent Ideally, the cosine distance between these two feature vectors should be 0, as they belong to the same identity.

{\bf 4. Reconstruction Loss}: To keep non-translating features from the input intact in the output image, we use a cyclic reconstruction loss \cite{CycleGAN} $L_{rec}$ between $I_{a}$ and its reconstruction $G(G(I_{a},f_b),f_a)$, computed as:
\vspace{-0.1cm}
\begin{equation}
L_{rec} = \mathbb{E}_{I_a,f_b,f_a}[\left \| I_a - G(G(I_a,f_b),f_a) \right \|_1]
\label{eq:L_rec}
\end{equation}

\begin{table*}[t]
\begin{center}
\caption{Quantitative comparison with popular GAN models on held out CMU-MultiPIE \cite{MultiPIE} test set.}
\begin{footnotesize}
\begin{tabular}{  | c | c| c| c| c| c| c| }
\hline
\begin{tabular}[x]{@{}c@{}}{\bf Models}\end{tabular} &
\begin{tabular}[x]{@{}c@{}}{\bf FID} \cite{FID} $\downarrow$\end{tabular} & \begin{tabular}[x]{@{}c@{}}{\bf LPIPS} \cite{ZhangCVPR18} $\downarrow$\end{tabular} &
\begin{tabular}[x]{@{}c@{}}{\bf SSIM} \cite{SSIM} $\uparrow$\end{tabular} & \begin{tabular}[x]{@{}c@{}}{\bf Match Score} \cite{ResNet,VGGFace2} $\uparrow$\end{tabular} & \begin{tabular}[x]{@{}c@{}}{\bf Quality Score} $\uparrow$ \end{tabular} & \begin{tabular}[x]{@{}c@{}}{\bf Human Preference} $\uparrow$ \end{tabular}\\

\hline
\hline
  \begin{tabular}[x]{@{}c@{}}{\bf StarGAN} \cite{StarGAN}\end{tabular} & 38.745 & 0.126 & 0.559 & 0.635 & 5.200 & 22.3\%\\
  \hline
    \begin{tabular}[x]{@{}c@{}}{\bf StarGAN w/ $L_{qual}$} \end{tabular} & 34.045 & 0.123 & 0.567 & 0.647 & 5.391 & 34.7\%\\
        \hline
      \begin{tabular}[x]{@{}c@{}}{\bf StarGAN-v2 \cite{StarGAN2}} \end{tabular} & 54.842 & 0.212 & 0.415 & 0.202 & 5.172 & 3.75\% \\
    \hline
      \begin{tabular}[x]{@{}c@{}}{\bf LEGAN} \end{tabular} & {\bf 29.964} & {\bf 0.120} & {\bf 0.649} & {\bf 0.649} & {\bf 5.853} & {\bf 39.3\%}\\
    \hline
    \hline
      \begin{tabular}[x]{@{}c@{}}{\bf Real Images} \end{tabular} & 12.931 & -  & - & 0.739 & 5.921 & -\\

      \hline
\end{tabular}
\label{Tab:Comp_Quant}
\end{footnotesize}
\end{center}
\vspace{-0.4cm}
\end{table*}

\begin{table*}[t]
\begin{center}
\caption{Quantitative performance of popular GAN models, trained only on CMU-MultiPIE \cite{MultiPIE}, when tested on AFLW \cite{aflw} images.}
\begin{footnotesize}
\begin{tabular}{  | c | c| c| c| c| c| }
\hline
\begin{tabular}[x]{@{}c@{}}{\bf Metrics}\end{tabular} & \begin{tabular}[x]{@{}c@{}}{\bf FID} \cite{FID} $\downarrow$\end{tabular} & \begin{tabular}[x]{@{}c@{}}{\bf LPIPS} \cite{ZhangCVPR18} $\downarrow$\end{tabular} &
\begin{tabular}[x]{@{}c@{}}{\bf SSIM} \cite{SSIM} $\uparrow$\end{tabular} & \begin{tabular}[x]{@{}c@{}}{\bf Match Score} \cite{ResNet,VGGFace2} $\uparrow$ \end{tabular} & \begin{tabular}[x]{@{}c@{}}{\bf Quality Score} $\uparrow$ \end{tabular}\\
\hline
\hline
 \begin{tabular}[x]{@{}c@{}}{\bf StarGAN} \cite{StarGAN}\end{tabular} & 43.987 & 0.302 & {\bf 0.639} & 0.622 & 6.41\\
  \hline
    \begin{tabular}[x]{@{}c@{}}{\bf StarGAN w/ $L_{qual}$} \end{tabular} & 41.641 & 0.268 & 0.624 & 0.614 & {\bf 6.73}\\
        \hline
      \begin{tabular}[x]{@{}c@{}}{\bf StarGAN-v2 \cite{StarGAN2}} \end{tabular} & 59.328 & 0.445 & 0.279 & 0.213 & 6.60\\
    \hline
      \begin{tabular}[x]{@{}c@{}}{\bf LEGAN} \end{tabular} & {\bf 38.794} & {\bf 0.271} & 0.622 & {\bf 0.628} & 6.68\\
    \hline
    \hline
      \begin{tabular}[x]{@{}c@{}}{\bf Real Images} \end{tabular} & - & - & - & - & 6.83\\
    \hline
\end{tabular}
\label{Tab:Comp_Quant_aflw}
\end{footnotesize}
\end{center}
\vspace{-0.8cm}
\end{table*}

\begin{table*}[t]
\begin{center}
\caption{Quantitative performance of popular GAN models, trained only on CMU-MultiPIE \cite{MultiPIE}, when tested on CelebA \cite{celebA} images.}
\begin{footnotesize}
\begin{tabular}{  | c | c| c| c| c| c| }
\hline
\begin{tabular}[x]{@{}c@{}}{\bf Metrics}\end{tabular} & \begin{tabular}[x]{@{}c@{}}{\bf FID} \cite{FID} $\downarrow$\end{tabular} & \begin{tabular}[x]{@{}c@{}}{\bf LPIPS} \cite{ZhangCVPR18} $\downarrow$\end{tabular} &
\begin{tabular}[x]{@{}c@{}}{\bf SSIM} \cite{SSIM} $\uparrow$\end{tabular} & \begin{tabular}[x]{@{}c@{}}{\bf Match Score} \cite{ResNet,VGGFace2} $\uparrow$ \end{tabular} & \begin{tabular}[x]{@{}c@{}}{\bf Quality Score} $\uparrow$ \end{tabular}\\
\hline
\hline
 \begin{tabular}[x]{@{}c@{}}{\bf StarGAN} \cite{StarGAN}\end{tabular} & 42.089 & 0.173 & 0.623 & 0.620 & 6.98\\
  \hline
    \begin{tabular}[x]{@{}c@{}}{\bf StarGAN w/ $L_{qual}$} \end{tabular} & 36.189 & {\bf 0.145} & 0.614 & 0.621 & {\bf 7.15} \\
        \hline
      \begin{tabular}[x]{@{}c@{}}{\bf StarGAN-v2 \cite{StarGAN2}} \end{tabular} & 50.360 & 0.311 & 0.312 & 0.295 & 6.89 \\
    \hline
      \begin{tabular}[x]{@{}c@{}}{\bf LEGAN} \end{tabular} & {\bf 29.059} & 0.184 & {\bf 0.662} & {\bf 0.635} & 7.07\\
    \hline
    \hline
      \begin{tabular}[x]{@{}c@{}}{\bf Real Images} \end{tabular} & - & - & - & - & 7.31\\
    \hline

\end{tabular}
\label{Tab:Comp_Quant_celeba}
\end{footnotesize}
\end{center}
\vspace{-0.4cm}
\end{table*}

{\bf 5. Quality Loss}: We use $Q$'s predictions for further improving with perceptual realism of the synthetic images. Masked versions of the input image ${I_{a}}'$, the synthesized output ${G(I_a,f_b)}'$ and the reconstructed input ${G(G(I_a,f_b),f_a)}'$, produced using facial landmarks extracted by \cite{BulatLandmark}, are used for loss computation as follows:
\begin{multline}
L_{qual} = \mathbb{E}_{I_a,f_b}[\left \| q - Q({G(I,f_b)}') \right \|_1] + \\ \mathbb{E}_{I_a,f_b,f_a}[\left \| Q({I_a}') - Q({G(G(I,f_b),f_a)}') \right \|_1]
\label{eq:L_qual}
\end{multline}
where $q$ is a hyper-parameter that can be tuned to lie between 5 (realistic) and 10 (hyper-realistic). We find synthesis results to be optimal when $q$ = 8.

{\bf Full Loss}: We also apply total variation loss \cite{Johnson} $L_{tv}$ on $G(I_a,f_b)$ and $G(G(I_a,f_b),f_a)$ to smooth boundary pixels and set the final training objective $L$ as a weighted sum of the six losses as:
\vspace{-0.1cm}
\begin{equation}
L = L_{adv} + \lambda_{1}L_{cls} + \lambda_{2}L_{rec} + \lambda_{3}L_{id} + \lambda_{4}L_{qual} + \lambda_{5}L_{tv}
\label{eq:L_full}
\vspace{-0.2cm}
\end{equation}

\begin{figure*}[t]
\centering
   \includegraphics[width=0.9\linewidth]{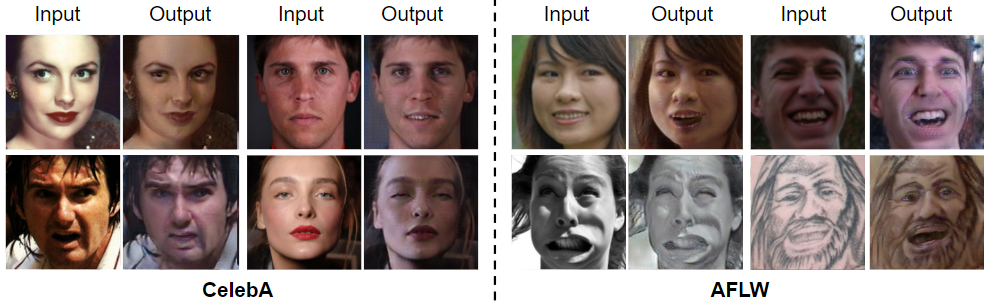}
   \caption{LEGAN synthesized output images from randomly sampled images from the AFLW \cite{aflw} and CelebA \cite{celebA} datasets. Although, trained only on frontal MultiPIE images, LEGAN can realistically manipulate lighting and expression in images with variance in pose, resolution and domain (sketch).}
\label{fig:collage}
\vspace{-0.4cm}
\end{figure*}

\begin{table*}
\begin{center}
\caption{Verification performance (TPR@FPR = 1\%) of LightCNN-29 \cite{LightCNN} on IJB-B \cite{IJBB} and LFW \cite{LFW} with and without LEGAN based augmentation.}
\begin{footnotesize}
\begin{tabular}{  | c | c| c| c| c| }
\hline
\begin{tabular}[x]{@{}c@{}}{\bf Training Data}\end{tabular} & \begin{tabular}[x]{@{}c@{}}{\bf Real Images} \cite{CASIA} (\# Identities)\end{tabular} & \begin{tabular}[x]{@{}c@{}}{\bf Synthetic Images} (\# Identities) \end{tabular} & \begin{tabular}[x]{@{}c@{}}{\bf IJB-B \cite{IJBB} Performance}\\\end{tabular} & \begin{tabular}[x]{@{}c@{}}{\bf LFW \cite{LFW} Performance}\end{tabular}\\
\hline
  \begin{tabular}[x]{@{}c@{}}Original\end{tabular}  & \begin{tabular}[x]{@{}c@{}}439,999 (10,575)\end{tabular} & 0 & 0.954 $\pm$ 0.002 & 0.966 $\pm$ 0.002 \\
  \hline
    \begin{tabular}[x]{@{}c@{}}Augmented\end{tabular}  & \begin{tabular}[x]{@{}c@{}}439,999 (10,575)\end{tabular} & \begin{tabular}[x]{@{}c@{}}439,999 (10,575)\end{tabular} & {\bf 0.967 $\pm$ 0.001} & {\bf 0.972 $\pm$ 0.001} \\
  \hline
\end{tabular}
\label{Tab:IJBB_CW}
\end{footnotesize}
\end{center}
\vspace{-0.4cm}
\end{table*}

\begin{table*}
\begin{center}
\caption{Model \cite{ertugrul} performance (ROC AUC) on AffectNet \cite{AffectNet} with and without LEGAN based augmentation}
\begin{footnotesize}
\begin{tabular}{ | c | c| c| c| c| c| c| }
\hline
\begin{tabular}[x]{@{}c@{}}{\bf Training Data}\end{tabular} & \begin{tabular}[x]{@{}c@{}}{\bf Real Images} \cite{AffectNet} \end{tabular} & \begin{tabular}[x]{@{}c@{}}{\bf Synthetic Images} \end{tabular} & \begin{tabular}[x]{@{}c@{}}{\bf `Neutral'}\end{tabular} & \begin{tabular}[x]{@{}c@{}}{\bf `Happy'}\end{tabular} & \begin{tabular}[x]{@{}c@{}}{\bf `Surprise'}\end{tabular} & \begin{tabular}[x]{@{}c@{}}{\bf`Disgust'}\end{tabular}\\
\hline
  \begin{tabular}[x]{@{}c@{}}Original\end{tabular}  & \begin{tabular}[x]{@{}c@{}}204,325 \end{tabular} & 0 & 0.851 $\pm$ 0.005 & 0.955 $\pm$ 0.001 & 0.873 $\pm$ 0.004 & 0.887 $\pm$ 0.005 \\
  \hline
    \begin{tabular}[x]{@{}c@{}}Augmented\end{tabular}  & \begin{tabular}[x]{@{}c@{}}204,325\end{tabular} & \begin{tabular}[x]{@{}c@{}}279,324\end{tabular} & {\bf 0.868 $\pm$ 0.005} & {\bf 0.956 $\pm$ 0.001} & {\bf 0.890 $\pm$ 0.003} & {\bf 0.897 $\pm$ 0.001} \\
  \hline
\end{tabular}
\label{Tab:AffectNet}
\end{footnotesize}
\end{center}
\vspace{-0.8cm}
\end{table*}

\section{Experiments and Results}
\label{sec:experiments}
\vspace{-0.2cm}
\noindent {\bf Training Data}. We utilize 36,657 frontal RGB images from the CMU-MultiPIE dataset \cite{MultiPIE}, with 20 different lighting conditions and 6 acted facial expressions, to build our model. For training we use 33,305 images of 303 subjects and the remaining 3,352 images of 34 subjects for testing. The training data is highly skewed towards `Neutral' and `Smile' compared to the other 4 expressions but the distribution is almost uniform for the lighting classes. We align each image using their eye landmarks extracted with \cite{BulatLandmark} and resize to 128$\times$128$\times$3. We do not fine-tune LEGAN on any other data and solely rely on its generalizability for the different experimental tasks.

\noindent {\bf Testing Data}. Along with MultiPIE's held out test set, we also utilize the AFLW \cite{aflw} and CelebA \cite{celebA} datasets to test the robustness of LEGAN towards in-the-wild conditions. We do not train or fine-tune LEGAN on these datasets and use the model trained on MultiPIE for translation tasks.

\noindent {\bf Augmentation: Recoloring}. Since MultiPIE \cite{MultiPIE} was acquired in a controlled setting, it has more or less uniform hue and saturation across all images. To artificially inject some diversity in the overall image color, and prevent visual overfitting, we build an image colorization model. Inspired by \cite{ColorZhang}, we train two separate pix2pix \cite{pix2pix} style GAN models where the input is a grayscale image and the target is set as its colored counterpart (\ie original image). To manipulate the color style of the same grayscale image differently, we train these two models with randomly sampled 10K images from the UMDFaces \cite{UMDFaces} and FFHQ \cite{ProgressiveGAN} datasets respectively. The trained generators are used to augment the color style of the MultiPIE training set.

\noindent {\bf Augmentation: White Balancing}. In addition to changing the color style, we also artificially edit the white balance of the training images by utilizing the pre-trained model from \cite{afifi2020deepWB}. This model can automatically correct the white balance of an image or render it with different camera presets. For each training input, we randomly use it as is or select its recolored or color corrected version, and pass it to LEGAN.

\noindent {\bf Implementation Details}. To learn the model, we use the Adam optimizer \cite{Adam} with a learning rate of 0.0001 and parameters $\beta_1$ and $\beta_2$ set to 0.5 and 0.999 respectively. The different loss weights $\lambda_{gp}$, $\lambda_{1}$, $\lambda_{2}$, $\lambda_{3}$ $\lambda_{4}$ and $\lambda_{5}$ are set empirically to 10, 20, 10, 10, 0.5 and 0.0000001 respectively. As done in \cite{GANimation}, we train $D$ 5 times for each training iteration of the $G$. We train the LightCNN-29 model \cite{LightCNN} on the CASIA-WebFace dataset \cite{CASIA} and utilize it as our identity network \emph{T}. Features from its penultimate layer are used to compute $L_{id}$. LEGAN is trained with a batch size of 10 on a single Tesla V100 GPU for 100 epochs.

\noindent {\bf Comparison with Other GAN Models}: Our proposed model is simple and unique as it does not require 3DMM information or external synthetic images during training \cite{DiscoFaceGAN,CONFIG} nor do we need to fine-tune our model during testing on input images. Due to this simplicity, we choose 2 popular publicly available unpaired domain translation models for comparison - StarGAN \cite{StarGAN} and the more recent StarGAN-v2 \cite{StarGAN2}. We train these models for lighting and expression manipulation with the same MultiPIE \cite{MultiPIE} training split for 100 epochs. Additionally, to gauge the effect of $L_{qual}$ on off-the-shelf models we train StarGAN separately with the auxiliary discriminator added (StarGAN w/ $L_{qual}$).

\noindent {\bf Metrics for Quality Estimation}. To evaluate the quality of synthetic face images generated by LEGAN and other GAN models, we compare the synthesized output with the corresponding target image\footnote{Since target images are not available for the AFLW \cite{aflw} and CelebA \cite{celebA} datasets, we evaluate metrics between the source and output images.} using these metrics - (1) {\bf FID} \cite{FID} and (2) {\bf LPIPS} \cite{ZhangCVPR18} to gauge the realism, (3) {\bf SSIM} \cite{SSIM} to measure noise, and (4) {\bf face match score} using pre-trained ResNet50 \cite{ResNet,VGGFace2} features and Pearson correlation coefficient. We also use our trained quality estimator to directly extract the (5) {\bf quality score} of real and synthetic images.

\noindent {\bf Human Evaluation}. We also run a perceptual study using face images generated by these models where we ask 17 non-expert human raters to pick an image from a lineup that best matches - (1) a target facial expression and (2) a target lighting condition, while (3) preserving the identity for 30 different MultiPIE \cite{MultiPIE} subjects. The raters are first shown real examples of the target expressions and lighting conditions. Each lineup consists of an actual image of the subject with bright lighting and neutral expression and the same subject synthesized for the target expression and lighting by the StarGAN \cite{StarGAN}, StarGAN w/ $L_{qual}$, StarGAN-v2 \cite{StarGAN2} and LEGAN, presented in a randomized order. We aggregate the rater votes across all rows and normalize them for each model (rightmost column of Table \ref{Tab:Comp_Quant}).

\noindent {\bf Quantitative Results}. As can be seen from Table \ref{Tab:Comp_Quant}, LEGAN synthesizes perceptually superior face images (FID, LPIPS, Quality Score) while retaining subject identity (Match Score) better than the other GAN models. As validated by the human evaluation, LEGAN also effectively translates the input image to the target lighting and expression conditions. Surprisingly, the StarGAN-v2 fails to generate realistic images in these experiments. This can be attributed to the fact that our task of joint lighting and expression manipulation presents the model with a much higher number of possible transformation domains (101 to be exact). The StarGAN-v2 model does learn to separate and transform lighting and expression to a certain degree but fails to decouple the other image attributes like identity, gender and race\footnote{We are not the first to encounter this issue, as shared here: \url{https://github.com/clovaai/stargan-v2/issues/21}}. Therefore, it synthesizes images that can be easily picked out by human perusal or identity matching. We also find adding $L_{qual}$ to StarGAN improves almost all its metric scores underpinning the value our quality estimator $Q$ even when coupled with off-the-shelf models. On top of enhancing the overall sharpness, $L_{qual}$ removes bullet-hole artifacts, similar to \cite{StyleGAN2}, from the peripheral regions of the output face. Such an artifact can be seen in (row 4, col 2, forehead-hair boundary) of Figure \ref{fig:Qual_Comp}, which is eliminated by adding $L_{qual}$ (row 4, col 3).

For the unseen, in-the-wild datasets, the performance of LEGAN is mostly superior to the other models, as shown in Tables \ref{Tab:Comp_Quant_aflw} and \ref{Tab:Comp_Quant_celeba} respectively. Due to the non-uniform nature of the data, especially the facial pose, most of the metrics deteriorate from Table \ref{Tab:Comp_Quant}. However, the boost in quality score overall suggests the high quality images from AFLW \cite{aflw} and especially CelebA \cite{celebA} to be visually more appealing than images from MultiPIE \cite{MultiPIE}. Some sample results have been shared in Figure \ref{fig:collage}.

\noindent {\bf Effectiveness as Training Data Augmenter}: We examine the use of LEGAN as training data augmenter for face verification and expression recognition tasks using the IJB-B \cite{IJBB}, LFW \cite{LFW} and AffectNet \cite{AffectNet} datasets. For face verification, we use the CASIA-WebFace dataset \cite{CASIA} and the LightCNN-29 \cite{LightCNN} architecture due to their popularity in this domain. We randomly sample 439,999 images of 10,575 subjects from \cite{CASIA} for training and 54,415 images for validation. We augment the training set by randomly manipulating the lighting and expression of each image (Table \ref{Tab:IJBB_CW}, row 2). The LightCNN-29 model is trained from scratch separately with the original and augmented sets and its weights saved when validation loss plateaus across epochs. These saved snapshots are then used to extract features from a still image or video frame in the IJB-B \cite{IJBB} dataset. For each IJB-B template, a mean feature is computed using video and media pooling operations \cite{masiFG17} and match score between such features is calculated with Pearson correlation. For the LFW \cite{LFW} images, we simply compare features between similar and dissimilar identities. To measure statistical significance of any performance benefit, we run each training 3 separate times. We find the model trained with the augmented data to improve upon the verification performance of the baseline (Table \ref{Tab:IJBB_CW}). This suggests that the LEGAN generated images retain their original identity and can boost the robustness of classification models towards intra-class variance in expressions and lighting.

For expression classification, we use a modified version of the AU-classification model from \cite{ertugrul} (Leaky ReLU \cite{LeakyRelu} and Dropout added) and manually annotated AffectNet \cite{AffectNet} images for the (`Neutral', `Happy', `Surprise', `Disgust') classes, as these 4 expressions overlap with MultiPIE \cite{MultiPIE}. The classification model is trained with 204,325 face images from AffectNet's training split, which is highly skewed towards the `Happy' class (59\%) and has very few images for the `Surprise' (6.2\%) and `Disgust' (1.6\%) expressions. To balance the training distribution, we populate each sparse class with synthetic images generated by LEGAN from real images belonging to any of the other 3 classes. We use the original and augmented (balanced) data separately to train two versions of the model for expression classification. As there is no test split, we use the 2,000 validation images for testing, as done in other works \cite{FECNet}. We find the synthetic images, when used in training, to substantially improve test performance especially for the previously under-represented `Surprise' and `Disgust' classes (Table \ref{Tab:AffectNet}). This further validates the realism of the expressions generated by LEGAN.

\vspace{-0.3cm}
\section{Conclusion}
\label{sec:conclusion}
\vspace{-0.2cm}
We propose LEGAN, a GAN framework for performing many-to-many joint manipulation of lighting and expressions of an existing face image without requiring paired training data. Instead of translating the image representations in an entangled feature space like \cite{StarGAN}, LEGAN estimates transformation maps in the decomposed lighting and expression sub-spaces before combining them to get the desired output image. To enhance the perceptual quality of the synthetic images, we directly integrate a quality estimation model into LEGAN's pipeline as an auxiliary discriminator. This quality estimation model, built with synthetic face images from different methods \cite{ProgressiveGAN,StyleGen,FaceFPP,DeepFake,SREFI2} and their crowd-sourced naturalness ratings, is trained using a margin based regression loss to capture the subjective nature of human judgement. The usefulness of LEGAN's feature disentangling towards synthesis quality is shown by objective comparison \cite{FID,ZhangCVPR18,SSIM} of its synthesized images to that of the other popular GAN models like StarGAN \cite{StarGAN} and StarGAN-v2 \cite{StarGAN2}. These experiments also highlight the usefulness of the proposed quality estimator in LEGAN and StarGAN w/ $L_{qual}$, specifically comparing the latter to vanilla StarGAN.

As a potential application, we use LEGAN as training data augmenter for face verification on the IJB-B \cite{IJBB} and LFW \cite{LFW} datasets and for facial expression classification on the AffectNet \cite{AffectNet} dataset (Tables \ref{Tab:IJBB_CW} and \ref{Tab:AffectNet}). An improvement in the verification scores in both datasets suggests LEGAN can enhance the intra-class variance while preserving subject identity. The boost in expression recognition performance validates the realism of the LEGAN generated facial expressions. The output quality, however, could be further improved when translating from an intense expression to another. We plan to address this by - (1) using attention masks in our encoder modules, and (2) building translation pathways of facial action units \cite{AUforExpression} while going from one expression to another. Another future goal is to incorporate a temporal component in LEGAN for synthesizing a sequence of coherent frames.

{\small
\bibliographystyle{ieee_fullname}
\bibliography{egbib}
}

\vspace{-1cm}
\section{Quality Estimation Model: Architecture Details}
\vspace{-0.1cm}
We share details of the architecture of our quality estimator $Q$ in Table \ref{Tab:Q}. The fully connected layers in $Q$ are denoted as `fc' while each convolution layer, represented as `conv', is followed by Leaky ReLU \cite{LeakyRelu} activation with a slope of 0.01.

\begin{table}[h]
\begin{center}
\caption{Detailed architecture of our quality estimation model $Q$ (input size is 128$\times$128$\times$3).}
\begin{small}
\begin{tabular}{  | c | c| c| }
\hline
\begin{tabular}[x]{@{}c@{}}{\bf Layer}\end{tabular} & \begin{tabular}[x]{@{}c@{}}{\bf Filter/Stride/Dilation}\end{tabular} & \begin{tabular}[x]{@{}c@{}}{\bf \# of filters}\end{tabular}\\
\hline
\hline
  input & 128$\times$128 & 3\\
\hline
\hline
  conv0 & 4$\times$4/2/1 & 64 \\
  conv1 & 4$\times$4/2/1 & 128 \\
  conv2 & 4$\times$4/2/1 & 256 \\
  conv3 & 4$\times$4/2/1 & 512 \\
  conv4 & 4$\times$4/2/1 & 1024 \\
  conv5 & 4$\times$4/2/1 & 2048 \\
  \hline
  \hline
  fc0 & - & 256 \\
  fc1 & - & 1 \\
  \hline
\end{tabular}
\label{Tab:Q}
\end{small}
\end{center}
\vspace{-0.5cm}
\end{table}

\begin{figure*}[t]
\centering
   \includegraphics[width=0.7\linewidth]{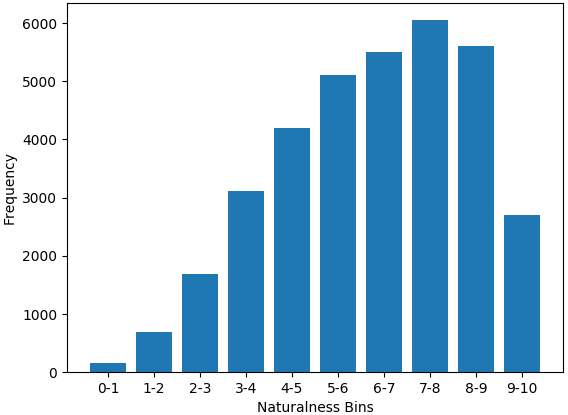}
   \caption{Histogram depicting the number of images in each naturalness bin, as rated by the Amazon Mechanical Turkers. Much more images fell on the `natural' half (5 - 10) rather than the `unnatural' one (0 - 5), suggesting the synthetic face images used in our study to be more or less realistic.}
\label{fig:AMT_Mean_Hist}
\vspace{-0.2cm}
\end{figure*}

\begin{table}
\begin{center}
\caption{Hourglass architecture for expression mask ($M_e$) synthesis in the generator $G$. The input size is 128$\times$128$\times$9, three RGB channels ($I_a$) and six expression channels ($c_e$).}
\begin{small}
\begin{tabular}{  | c | c| c| }
\hline
\begin{tabular}[x]{@{}c@{}}{\bf Layer}\end{tabular} & \begin{tabular}[x]{@{}c@{}}{\bf Filter/Stride/Dilation}\end{tabular} & \begin{tabular}[x]{@{}c@{}}{\bf \# of filters}\end{tabular}\\
\hline
\hline
  input & 128$\times$128/-/- & 9\\
\hline
\hline
  conv0 & 7$\times$7/1/1 & 64 \\
    \hline
  \hline
  conv1 & 4$\times$4/2/1 & 128 \\
  conv2 & 4$\times$4/2/1 & 256 \\
  \hline
  \hline
  RB0 & 3$\times$3/1/1 & 256\\
  RB1 & 3$\times$3/1/1 & 256\\
  RB2 & 3$\times$3/1/1 & 256\\
  RB3 & 3$\times$3/1/1 & 256\\
  RB4 & 3$\times$3/1/1 & 256\\
  RB5 & 3$\times$3/1/1 & 256\\
  \hline
  \hline
  PS0 & - & 256\\
  conv3 & 4$\times$4/1/1 & 128 \\
  PS1 & - & 128\\
  conv4 & 4$\times$4/1/1 & 64 \\
  \hline
  \hline
  conv5 ($M_e$) & 7$\times$7/1/1 & 3 \\
  \hline
\end{tabular}
\label{Tab:HG1}
\end{small}
\end{center}
\vspace{-0.5cm}
\end{table}

\begin{table}
\begin{center}
\caption{Hourglass architecture for lighting mask ($M_l$) synthesis in the generator $G$. The input size is 128$\times$128$\times$23, three RGB channels ($I_a$) and twenty expression channels ($c_l$).}
\begin{small}
\begin{tabular}{  | c | c| c| }
\hline
\begin{tabular}[x]{@{}c@{}}{\bf Layer}\end{tabular} & \begin{tabular}[x]{@{}c@{}}{\bf Filter/Stride/Dilation}\end{tabular} & \begin{tabular}[x]{@{}c@{}}{\bf \# of filters}\end{tabular}\\
\hline
\hline
  input & 128$\times$128/-/- & 23\\
\hline
\hline
  conv0 & 7$\times$7/1/1 & 64 \\
    \hline
  \hline
  conv1 & 4$\times$4/2/1 & 128 \\
  conv2 & 4$\times$4/2/1 & 256 \\
  \hline
  \hline
  RB0 & 3$\times$3/1/1 & 256\\
  RB1 & 3$\times$3/1/1 & 256\\
  RB2 & 3$\times$3/1/1 & 256\\
  RB3 & 3$\times$3/1/1 & 256\\
  RB4 & 3$\times$3/1/1 & 256\\
  RB5 & 3$\times$3/1/1 & 256\\
  \hline
  \hline
  PS0 & - & 256\\
  conv3 & 4$\times$4/1/1 & 128 \\
  PS1 & - & 128\\
  conv4 & 4$\times$4/1/1 & 64 \\
  \hline
  \hline
  conv5 ($M_l$) & 7$\times$7/1/1 & 3 \\
  \hline
\end{tabular}
\label{Tab:HG2}
\end{small}
\end{center}
\vspace{-0.5cm}
\end{table}

\begin{table}[h]
\begin{center}
\caption{Hourglass architecture for target image ($G(I_a,f_b)$) synthesis in the generator $G$. The input size is 128$\times$128$\times$6, three expression mask channels ($M_e$) and three lighting mask channels ($M_l$).}
\begin{small}
\begin{tabular}{  | c | c| c| }
\hline
\begin{tabular}[x]{@{}c@{}}{\bf Layer}\end{tabular} & \begin{tabular}[x]{@{}c@{}}{\bf Filter/Stride/Dilation}\end{tabular} & \begin{tabular}[x]{@{}c@{}}{\bf \# of filters}\end{tabular}\\
\hline
\hline
  input & 128$\times$128/-/- & 6\\
\hline
\hline
  conv0 & 7$\times$7/1/1 & 64 \\
    \hline
  \hline
  conv1 & 4$\times$4/2/1 & 128 \\
  conv2 & 4$\times$4/2/1 & 256 \\
  \hline
  \hline
  RB0 & 3$\times$3/1/1 & 256\\
  RB1 & 3$\times$3/1/1 & 256\\
  RB2 & 3$\times$3/1/1 & 256\\
  RB3 & 3$\times$3/1/1 & 256\\
  RB4 & 3$\times$3/1/1 & 256\\
  RB5 & 3$\times$3/1/1 & 256\\
  \hline
  \hline
  PS0 & - & 256\\
  conv3 & 4$\times$4/1/1 & 128 \\
  PS1 & - & 128\\
  conv4 & 4$\times$4/1/1 & 64 \\
  \hline
  \hline
  conv5 ($G(I_a,f_b)$) & 7$\times$7/1/1 & 3 \\
  \hline
\end{tabular}
\label{Tab:HG3}
\end{small}
\end{center}
\vspace{-0.5cm}
\end{table}

\begin{table}[h]
\begin{center}
\caption{Detailed architecture of LEGAN's discriminator $D$ (input size is 128$\times$128$\times$3).}
\begin{small}
\begin{tabular}{  | c | c| c| }
\hline
\begin{tabular}[x]{@{}c@{}}{\bf Layer}\end{tabular} & \begin{tabular}[x]{@{}c@{}}{\bf Filter/Stride/Dilation}\end{tabular} & \begin{tabular}[x]{@{}c@{}}{\bf \# of filters}\end{tabular}\\
\hline
\hline
  input & 128$\times$128 & 3\\
\hline
\hline
  conv0 & 4$\times$4/2/1 & 64 \\
  conv1 & 4$\times$4/2/1 & 128 \\
  conv2 & 4$\times$4/2/1 & 256 \\
  conv3 & 4$\times$4/2/1 & 512 \\
  conv4 & 4$\times$4/2/1 & 1024 \\
  conv5 & 4$\times$4/2/1 & 2048 \\
  \hline
  \hline
  conv6 ($D_{src}$) & 3$\times$3/1/1 & 1 \\
  conv7 ($D_{cls}$) & 1$\times$1/1/1 & 26 \\
  \hline
\end{tabular}
\label{Tab:D}
\end{small}
\end{center}
\vspace{-0.5cm}
\end{table}

\section{Quality Estimation Model: Naturalness Rating Distribution in Training}
\vspace{-0.1cm}
In this section, we share the distribution of the naturalness ratings that we collected from the Amazon Mechanical Turk (AMT) experiment (Stage II). To do this, we average the perceptual rating for each synthetic face image from its three scores and increment the count of a particular bin in [(0 - 1), (1 - 2), ... , (8 - 9), (9 - 10)] based on the mean score. As described in Section 3 of the main paper, we design the AMT task such that a mean rating between 0 and 5 suggests the synthetic image to look `unnatural' while a score between 5 and 10 advocates for its naturalness. As can be seen in Figure \ref{fig:AMT_Mean_Hist}, majority of the synthetic images used in our study generates a mean score that falls on the `natural' side, validating their realism. When used to train our quality estimation model $Q$, these images tune its weights to look for the same perceptual features in other images while rating their naturalness.

\begin{figure*}[t]
\centering
   \includegraphics[width=0.7\linewidth]{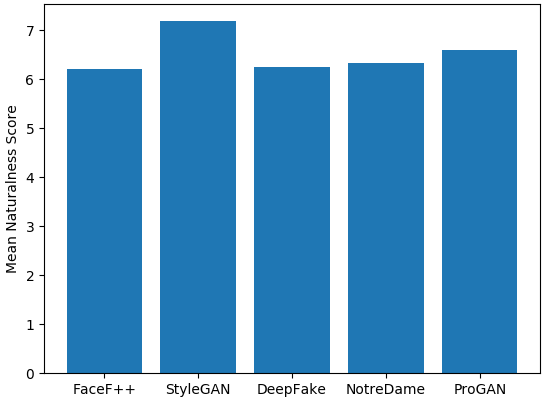}
   \caption{Mean naturalness rating of the different synthesis approaches used in our study \cite{ProgressiveGAN,StyleGen,DeepFake,FaceFPP,SREFI2}. As expected, the StyleGAN \cite{StyleGen} images are rated higher than others as they were pre-filtered for quality\cite{FilteredStyleGAN}.}
\label{fig:AMT_Mean_Hist_ByMethod}
\vspace{-0.1cm}
\end{figure*}

\begin{figure*}[t]
\centering
   \includegraphics[width=0.7\linewidth]{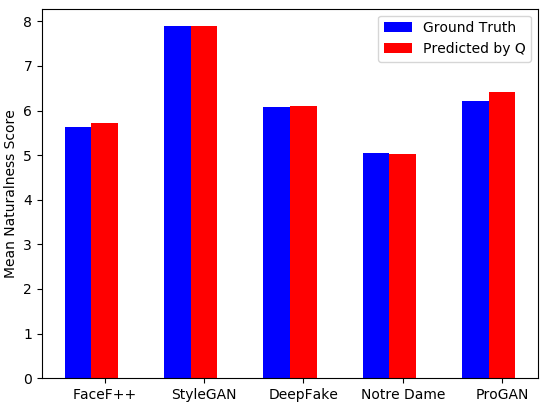}
   \caption{Mean naturalness rating, as estimated by Turkers (blue) and predicted by our trained quality estimation model $Q$ (red), for the different synthesis approaches used in our study \cite{ProgressiveGAN,StyleGen,DeepFake,FaceFPP,SREFI2}. These ratings are specifically for images from the test split in our experiments, so $Q$ never encountered them during training. Yet, $Q$ is able to predict the naturalness of these images with a high degree of certainty.}
\label{fig:Qual_Test}
\vspace{-0.5cm}
\end{figure*}

\begin{figure*}[t]
\centering
   \includegraphics[width=1.0\linewidth]{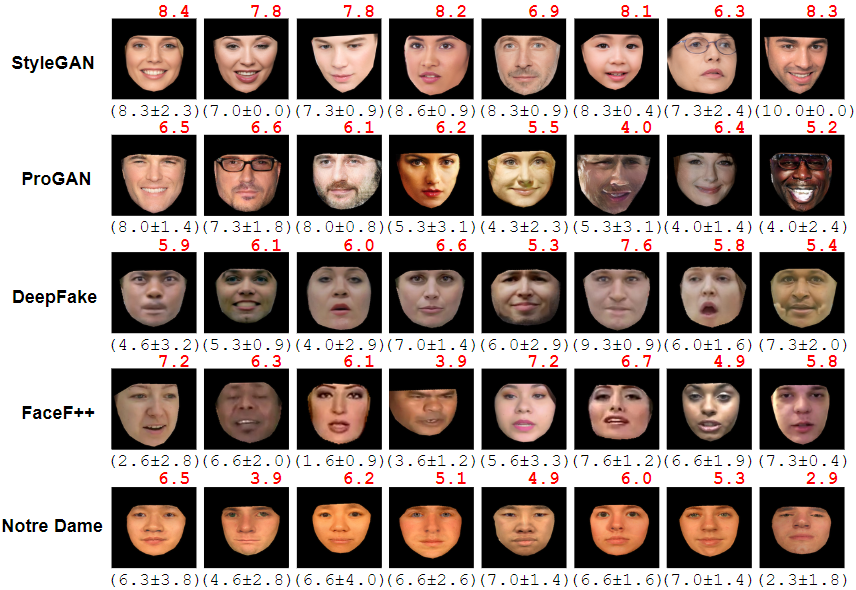}
   \caption{Perceptual quality predictions by our trained quality estimation model ($Q$) on sample test images generated using \cite{StyleGen,ProgressiveGAN,DeepFake,FaceFPP,SREFI2}. For each image, the (mean $\pm$ standard deviation) of the three naturalness scores, collected from AMT, is shown below while $Q$'s prediction is shown above in red.}
\label{fig:AMT_Gallery}
\vspace{-0.25cm}
\end{figure*}

To further check the overall perceptual quality of each of the different synthesis approaches used in our study \cite{ProgressiveGAN,StyleGen,DeepFake,FaceFPP,SREFI2}, we separately find the mean rating for each synthetic face image generated by that method, depicted in Figure \ref{fig:AMT_Mean_Hist_ByMethod}. It comes as no surprise for the StyleGAN \cite{StyleGen} images to rank the highest, with a mean score over 7, as its face images were pre-filtered for quality \cite{FilteredStyleGAN}. The other four approaches perform roughly the same, generating a mean score that falls between 6 and 7.

\section{Quality Estimation Model: Loss Function}
\vspace{-0.1cm}
Our loss uses the L2 norm between the predicted quality ($p$) and mean label ($\mu$) and then computes a second L2 norm between this distance and the standard deviation ($\sigma$). $\sigma$ acts as a margin in this case. If we consider $\mu$ as the center of a circle with radius of $\sigma$, then our loss tries to push $p$ towards the boundary to fully capture the subjectiveness of human perception. We also tried a hinge version of this loss:
$\max\left (0, \left ( \left \| \mu - p \right \|_{2}^{2} - \sigma \right )\right )$. This function penalizes $p$ falling outside the permissible circle while allowing it to lie anywhere within it. When $\sigma$ is low, both functions act similarly. We found the quality estimation model's ($Q$) predictions to be less stochastic when trained with the margin loss than the hinge. On a held-out test set, both losses performed similarly with only 0.2\% difference in regression accuracy. Experimental results with LEGAN, and especially StarGAN trained using $Q$ (Tables 1, 2, 3 in the main text), underpin the efficiency of the margin loss in comprehending naturalness. The improvements in perceptual quality, as demonstrated by LPIPS and FID, further justify its validity as a good objective for training $Q$.

\section{Quality Estimation Model: Prediction Accuracy During Testing}
\vspace{-0.1cm}
As discussed in Section 3 of the main text, we hold out 10\% of the crowd-sourced data (3,727 face images) for testing our quality estimation model $Q$ post training. Since $Q$ never encountered these images during training, we use them to evaluate the effectiveness of our model. We separately compute the mean naturalness score for each synthesis approach used in our study and compare this value with the average quality score as predicted by $Q$. The results can be seen in Figure \ref{fig:Qual_Test}. Overall, our model predicts the naturalness score for each synthesis method with a high degree of certainty. Some qualitative results can also be seen in Figure \ref{fig:AMT_Gallery}.

\section{LEGAN: Detailed Architecture}
\vspace{-0.1cm}
In this section, we list the different layers in the generator $G$ and discriminator $D$ of LEGAN. Since $G$ is composed of three hourglass networks, we separately describe their architecture in Tables \ref{Tab:HG1}, \ref{Tab:HG2} and \ref{Tab:HG3} respectively. The convolution layers, residual blocks and pixel shuffling layers are indicated as `conv', `RB', and `PS' respectively in the tables. After each of `conv' and `PS' layer in an hourglass, we use \emph{ReLU} activation and instance normalization \cite{InstanceNorm}, except for the last `conv' layer where a \emph{tanh} activation is used \cite{DCGAN,salimans}. The description of $D$ can be found in Table \ref{Tab:D}. Similar to $Q$, each convolution layer is followed by Leaky ReLU \cite{LeakyRelu} activation with a slope of 0.01 in $D$, except for the final two convolution layers that output the realness matrix $D_{src}$ and the classification map $D_{cls}$.

\begin{figure*}[t]
\centering
   \includegraphics[width=1.0\linewidth]{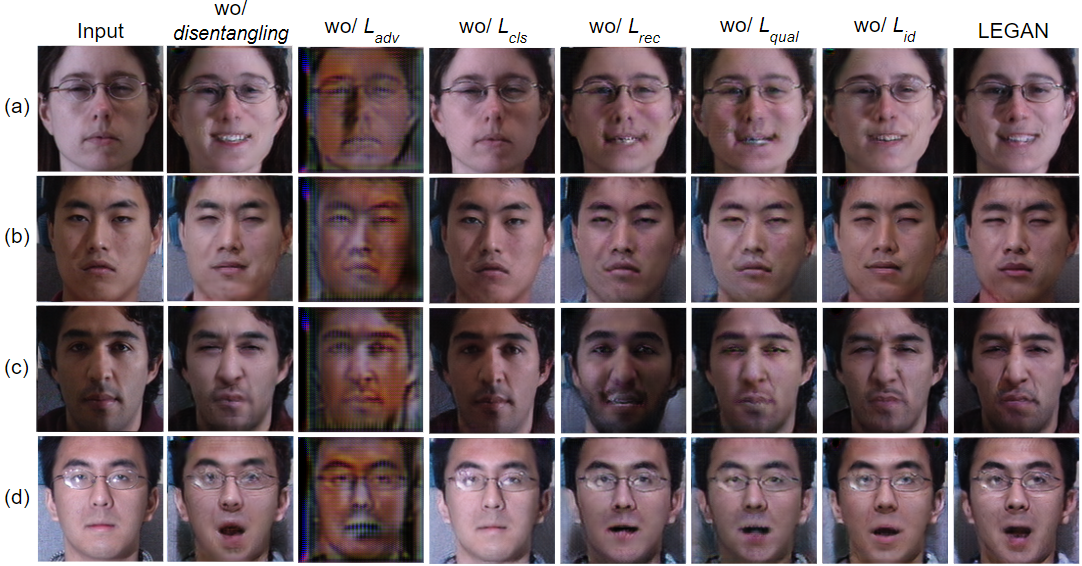}
   \caption{Sample qualitative results from LEGAN and its ablated variants on randomly sampled input images from MultiPIE \cite{MultiPIE} test set. The target expression and lighting conditions for each row are - (a) (Smile, Left Shadow), (b) (Squint, Ambient), (c) (Disgust, Left Shadow), and (d) (Surprise, Ambient).}
\label{fig:Qual_Ab}
\vspace{-0.1cm}
\end{figure*}

\begin{table*}[t]
\begin{center}
\caption{Ablation studies - quantitative results on held out CMU-MultiPIE \cite{MultiPIE} test set.}
\begin{footnotesize}
\begin{tabular}{  | c | c| c| c| c| c| }
\hline
\begin{tabular}[x]{@{}c@{}}{\bf Models}\end{tabular} & \begin{tabular}[x]{@{}c@{}}{\bf FID} \cite{FID} $\downarrow$\end{tabular} & \begin{tabular}[x]{@{}c@{}}{\bf LPIPS} \cite{ZhangCVPR18} $\downarrow$\end{tabular} &
\begin{tabular}[x]{@{}c@{}}{\bf SSIM} \cite{SSIM} $\uparrow$\end{tabular} & \begin{tabular}[x]{@{}c@{}}{\bf Match Score} \cite{ResNet,VGGFace2} $\uparrow$ \end{tabular} & \begin{tabular}[x]{@{}c@{}}{\bf Quality Score} $\uparrow$ \end{tabular}\\
\hline
\hline
 \begin{tabular}[x]{@{}c@{}}{\bf wo/ disentangling}\end{tabular} & 40.244 & 0.148 & 0.557 & 0.601 & 5.348\\
  \hline
 \begin{tabular}[x]{@{}c@{}}{\bf wo/ $L_{adv}$}\end{tabular} & 351.511 & 0.460 & 0.352 & 0.476 & 1.74\\
\hline
 \begin{tabular}[x]{@{}c@{}}{\bf wo/ $L_{cls}$}\end{tabular} & 30.236 & 0.139 & 0.425 & {\bf 0.717} & {\bf 5.873}\\
    \hline
 \begin{tabular}[x]{@{}c@{}}{\bf wo/ $L_{rec}$}\end{tabular} & 40.479 & 0.135 & 0.550 & 0.676 & 5.475\\
    \hline
 \begin{tabular}[x]{@{}c@{}}{\bf wo/ $L_{qual}$}\end{tabular} & 46.420 & 0.168 & 0.544 & 0.621 & 5.190\\
 \hline
  \begin{tabular}[x]{@{}c@{}}{\bf wo/ $L_{id}$}\end{tabular} & 35.429 & 0.140 & 0.566 & 0.587 & 5.861\\
    \hline
      \begin{tabular}[x]{@{}c@{}}{\bf LEGAN} \end{tabular} & {\bf 29.964} & {\bf 0.120} & {\bf 0.649} & 0.649 & 5.853\\
    \hline
\end{tabular}
\label{Tab:Ablation_Quant}
\end{footnotesize}
\end{center}
\vspace{-0.2cm}
\end{table*}

\section{LEGAN: Ablation Study}
To analyze the contribution of each loss component on synthesis quality, we prepare 5 different versions of LEGAN by removing (feature disentanglement, $L_{adv}$, $L_{cls}$, $L_{rec}$, $L_{qual}$, and $L_{id}$) from $G$ while keeping everything else the same. The qualitative and quantitative results, produced using MultiPIE \cite{MultiPIE} test data, are shown in Figure \ref{fig:Qual_Ab} and Table \ref{Tab:Ablation_Quant} respectively. For the quantitative results, the output image is compared with the corresponding target image in MultiPIE, and not the source image (\ie input).

As expected, we find $L_{adv}$ to be crucial for realistic hallucinations, in absence of which the model generates non-translated images totally outside the manifold of real images. The disentanglement of the lighting and expression via LEGAN's hourglass pair allows the model to independently generate transformation masks which in turn synthesize more realistic hallucinations. Without the disentanglement, the model synthesizes face images with pale-ish skin color and suppressed expressions. When $L_{cls}$ is removed, LEGAN outputs the input image back as the target attributes are not checked by $D$ anymore. Since the input image is returned back by the model, it generates a high face matching and mean quality score (Table \ref{Tab:Ablation_Quant}, third row). When the reconstruction error $L_{rec}$ is plugged off the output images lie somewhere in the middle, between the input and target expressions, suggesting the contribution of the loss in smooth translation of the pixels. Removing $L_{qual}$ and $L_{id}$ deteriorates the overall naturalness, with artifacts manifesting in the eye and mouth regions. As expected, the overall best metrics are obtained when the full LEGAN model with all the loss components is utilized.

\section{LEGAN: Optimal Upsampling}
\vspace{-0.1cm}
To check the effect of the different upsampling approaches on hallucination quality, we separately apply bilinear interpolation, transposed convolution \cite{Deconv} and pixel shuffling \cite{pixshuff} on the decoder module of the three hourglass networks in LEGAN's generator $G$. While the upsampled pixels are interpolated based on the original pixel in the first approach, the other two approaches explicitly learn the possible intensity during upsampling. More specifically, pixel shuffling blocks learn the intensity for the pixels in the fractional indices of the original image (i.e. the upsampled indices) by using a set convolution channels and have been shown to generate sharper results than transposed convolutions. Unsurprisingly, it generates the best quantitative results by outperforming the other two upsampling approaches on 3 out of the 5 objective metrics, as shown in Table \ref{Tab:Ablation_Up}. Hence we use pixel shuffling blocks in our final implementation of LEGAN.

\begin{figure*}[t]
\centering
   \includegraphics[width=1.0\linewidth]{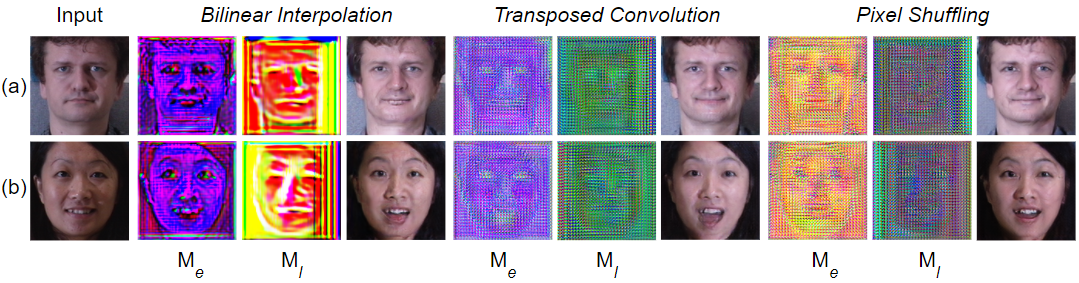}
   \caption{Adding different upsampling techniques in our decoder modules generates hallucinations with slightly different perceptual scores for the same input. Here the target expression and lighting conditions are set as - (a) (Smile, Bright), and (b) (Surprise, Left Shadow). However, the transformation masks $M_{e}$ and $M_{l}$ are smoother and more meaningful when bilinear interpolation is used for upsampling. Since both transposed convolution \cite{Deconv} and pixel shuffling \cite{pixshuff} learn the intensity of the upsampled pixels instead of simple interpolation, the masks they generate are more fragmented and discrete. We use pixel shuffling in our final LEGAN model.}
\label{fig:Up_Gallery}
\vspace{-0.1cm}
\end{figure*}

However, as can be seen in Figure \ref{fig:Up_Gallery}, the expression and lighting transformation masks $M_{e}$ and $M_{l}$ are more meaningful when interpolated rather than explicitly learned. This interpolation leads to a smoother flow of upsampled pixels with facial features and their transformations visibly more noticeable compared to transposed convolutions and pixel shuffling.

\begin{table*}[t]
\begin{center}
\caption{Effects of different upsampling - quantitative results on held out CMU-MultiPIE \cite{MultiPIE} test set.}
\begin{footnotesize}
\begin{tabular}{  | c | c| c| c| c| c| }
\hline
\begin{tabular}[x]{@{}c@{}}{\bf Models}\end{tabular} & \begin{tabular}[x]{@{}c@{}}{\bf FID} \cite{FID} $\downarrow$\end{tabular} & \begin{tabular}[x]{@{}c@{}}{\bf LPIPS} \cite{ZhangCVPR18} $\downarrow$\end{tabular} &
\begin{tabular}[x]{@{}c@{}}{\bf SSIM} \cite{SSIM} $\uparrow$\end{tabular} & \begin{tabular}[x]{@{}c@{}}{\bf Match Score} \cite{ResNet,VGGFace2} $\uparrow$ \end{tabular} & \begin{tabular}[x]{@{}c@{}}{\bf Quality Score} $\uparrow$ \end{tabular}\\
\hline
\hline
 \begin{tabular}[x]{@{}c@{}}{\bf Bilinear Interpolation}\end{tabular} & 29.933 & 0.128 & 0.630 & {\bf 0.653} & 5.823\\
  \hline
 \begin{tabular}[x]{@{}c@{}}{\bf Transposed Convolution \cite{Deconv}}\end{tabular} & {\bf 28.585} & 0.125 & 0.635 & 0.644 & 5.835\\
    \hline
  \begin{tabular}[x]{@{}c@{}}{\bf Pixel Shuffling \cite{pixshuff}} \end{tabular} & 29.964 & {\bf 0.120} & {\bf 0.649} & 0.649 & {\bf 5.853}\\
    \hline
\end{tabular}
\label{Tab:Ablation_Up}
\end{footnotesize}
\end{center}
\vspace{-0.2cm}
\end{table*}

\section{LEGAN: Optimal Value of $q$}
\begin{figure*}[t]
\centering
  \includegraphics[width=1.0\linewidth]{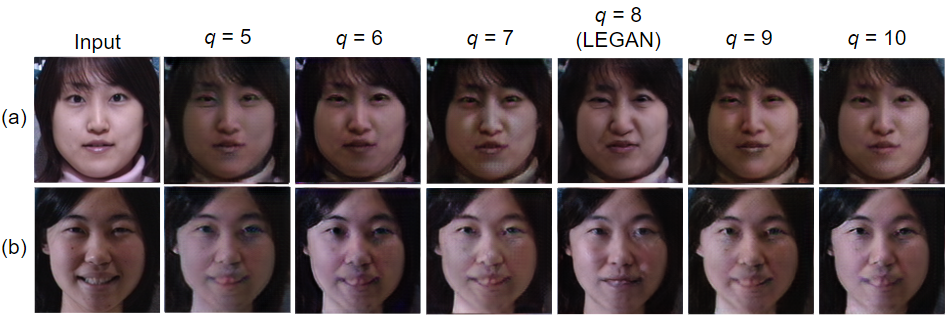}
  \caption{Sample results illustrating the effect of the hyper-parameter $q$ on synthesis quality. The input images are randomly sampled from the MultiPIE \cite{MultiPIE} test set with target expression and lighting conditions set as - (a) (Disgust, Left Shadow), and (b) (Neutral, Right Shadow). Since it generates more stable and noticeable expressions with fewer artifacts, we set $q$ = 8 for the final LEGAN model.}
\label{fig:Qual_q}
\vspace{-0.1cm}
\end{figure*}

\vspace{-0.1cm}
As discussed in the main text, we set the value of the hyper-parameter $q$ = 8 for computing the quality loss $L_{qual}$. We arrive at this specific value after experimenting with different possible values. Since $q$ acts as a target for perceptual quality while estimating $L_{qual}$ during the forward pass, it can typically range from 5 (realistic) to 10 (hyper-realistic). We set $q$ to all possible integral values between 5 and 10 for evaluating the synthesis results both qualitatively (Figure \ref{fig:Qual_q}) and quantitatively (Table \ref{Tab:Quant_q}).

As can be seen, when $q$ is set to 8, LEGAN generates more stable images with much less artifacts compared to other values of $q$. Also, the synthesized expressions are visibly more noticeable for this value of $q$ (Figure \ref{fig:Qual_q}, top row). When evaluated quantitatively, images generated by LEGAN with $q$ = 8 garner the best score for 4 out of 5 objective metrics. This is interesting as setting $q$ = 10 (and not 8) should ideally generate hyper-realistic images and consequently produce the best quantitative scores. We attribute this behavior of LEGAN to the naturalness distribution of the images used to train our quality model $Q$. Since majority of these images fell in the (7-8) and (8-9) bins, and very few in (9-10) (as shown in Figure \ref{fig:AMT_Mean_Hist}), $Q$'s representations are aligned to this target. As a result, $Q$ tends to rate hyper-realistic face images (i.e. images with mean naturalness rating between 8 - 10) with a score around 8. Such an example can be seen in the rightmost column of the first row in Figure \ref{fig:AMT_Gallery}, where $Q$ rates a hyper-realistic StyleGAN generated image \cite{StyleGen} as 8.3. Thus, setting $q$ = 8 for $L_{qual}$ computation (using trained $Q$'s weights) during LEGAN training produces the optimal results.

\begin{table*}[t]
\begin{center}
\caption{Quantitative results on the held out CMU-MultiPIE \cite{MultiPIE} test set by varying the value of the hyper-parameter $q$.}
\begin{footnotesize}
\begin{tabular}{  | c | c| c| c| c| c| }
\hline
\begin{tabular}[x]{@{}c@{}}{\bf Models}\end{tabular} & \begin{tabular}[x]{@{}c@{}}{\bf FID} \cite{FID} $\downarrow$\end{tabular} & \begin{tabular}[x]{@{}c@{}}{\bf LPIPS} \cite{ZhangCVPR18} $\downarrow$\end{tabular} &
\begin{tabular}[x]{@{}c@{}}{\bf SSIM} \cite{SSIM} $\uparrow$\end{tabular} & \begin{tabular}[x]{@{}c@{}}{\bf Match Score} \cite{ResNet,VGGFace2} $\uparrow$ \end{tabular} & \begin{tabular}[x]{@{}c@{}}{\bf Quality Score} $\uparrow$ \end{tabular}\\
\hline
\hline
 \begin{tabular}[x]{@{}c@{}}{\bf $q$ = 5}\end{tabular} & 41.275 & 0.143 & 0.550 & 0.642 & 5.337\\
  \hline
 \begin{tabular}[x]{@{}c@{}}{\bf $q$ = 6}\end{tabular} & 44.566 & 0.139 & 0.542 & 0.651 & 5.338\\
\hline
 \begin{tabular}[x]{@{}c@{}}{\bf $q$ = 7}\end{tabular} & 38.684 & 0.137 & 0.631 & {\bf 0.663} & 5.585\\
    \hline
 \begin{tabular}[x]{@{}c@{}}{\bf $q$ = 8 (LEGAN)}\end{tabular} & {\bf 29.964} & {\bf 0.120} & {\bf 0.649} & 0.649 & {\bf 5.853}\\
    \hline
 \begin{tabular}[x]{@{}c@{}}{\bf $q$ = 9}\end{tabular} & 42.772 & 0.137 & 0.637 & 0.659 & 5.686\\
 \hline
  \begin{tabular}[x]{@{}c@{}}{\bf $q$ = 10}\end{tabular} & 46.467 & 0.132 & 0.586 & 0.583 & 5.711\\
    \hline
\end{tabular}
\label{Tab:Quant_q}
\end{footnotesize}
\end{center}
\vspace{-0.2cm}
\end{table*}

\begin{figure}
\centering
  \includegraphics[width=1.0\linewidth]{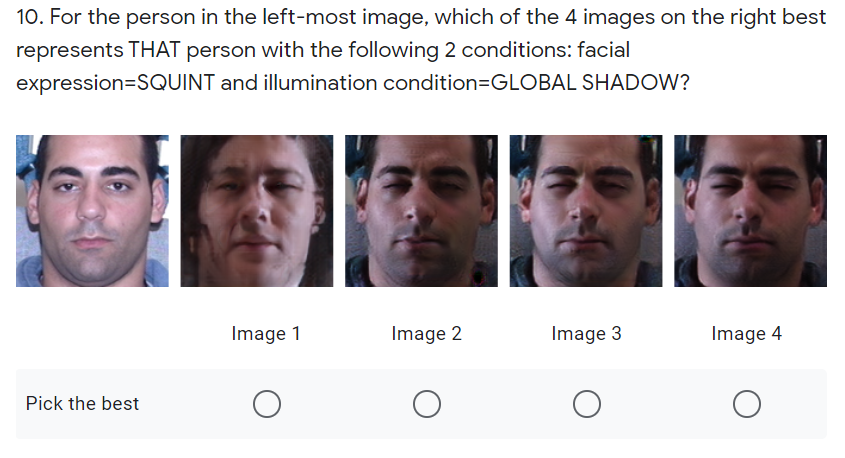}
  \caption{Our perceptual study interface: given a base face image with neutral expression and bright lighting (leftmost image), a rater is asked to select the image that best matches the target expression (`Squint') and lighting (`Right Shadow') for the same subject.}
\label{fig:Interface}
\end{figure}

\section{LEGAN: Perceptual Study Details}
In this section, we share more details about the interface used for our perceptual study. As shown in Figure \ref{fig:Interface}, we ask the raters to pick the image that best matches a target expression and lighting condition. To provide a basis for making judgement, we also share a real image of the same subject with neutral expression and bright lighting condition. However, this is not necessarily the input to the synthesis models for the target expression and lighting generation, as we want to estimate how these models do when the input image has more extreme expressions and lighting conditions. The image order is also randomized to eliminate any bias.

\section{LEGAN: Model Limitations}
\begin{figure*}
\centering
  \includegraphics[width=1.0\linewidth]{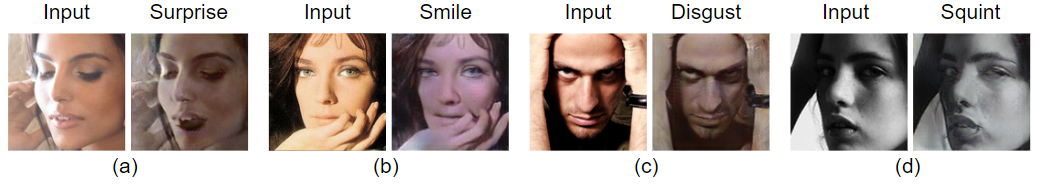}
  \caption{Failure cases: for each input image LEGAN fails to correctly generate the target facial expression. In (a) LEGAN manages to generate a surprised mouth but fails to open the subject's eyes, (b) the smile is half generated due to occlusion by the subject's fingers, (c) the target disgusted expression is missing, and (d) the subject's eyes are not squinted. Most of these failure cases are either due to non-frontal facial pose or occlusion.}
\label{fig:Fail}
\end{figure*}

Although LEGAN is trained on just frontal face images acquired in a controlled setting, it can still generate realistic new views even for non-frontal images with a variety of expressions, as shown in Figures \ref{fig:Gal_1} and \ref{fig:Gal_2}. However, as with any synthesis model, LEGAN also has its limitations. In majority of the cases where LEGAN fails to synthesize a realistic image, the input expression is irregular with non-frontal head pose or occlusion, as can be seen in Figure \ref{fig:Fail}. As a result, LEGAN fails to generalize and synthesizes images with incomplete translations or very little pixel manipulations. One way to mitigate this problem is to extend both our quality model and LEGAN to non-frontal facial poses and occlusions by introducing randomly posed face images during training.

\section{LEGAN: More Qualitative Results}
\begin{figure*}
\centering
  \includegraphics[width=0.9\linewidth]{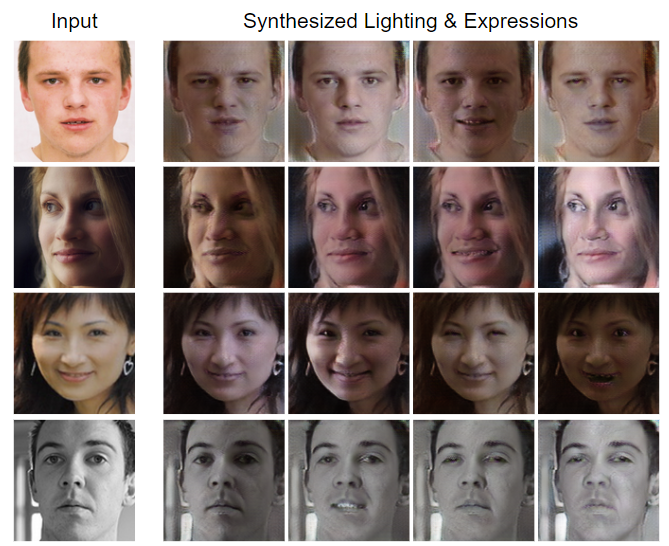}
  \caption{Synthesized expressions and lighting conditions for the same input image, as generated by LEGAN. These input images are randomly selected from the AFLW \cite{aflw} dataset and the results are generated by randomly setting different expression and lighting targets. LEGAN is trained on 33k frontal face images from MultiPIE \cite{MultiPIE} and we do not fine-tune the model on any other dataset. All images are 128$\times$128$\times$3.}
\label{fig:Gal_1}
\end{figure*}

\begin{figure*}
\centering
  \includegraphics[width=0.9\linewidth]{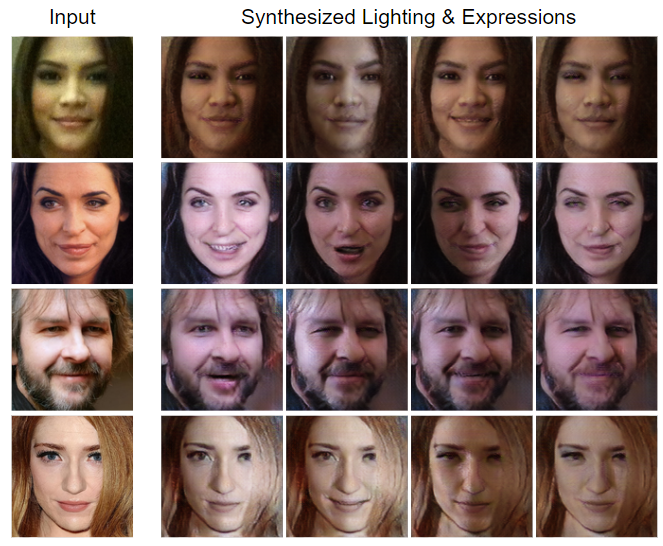}
  \caption{Synthesized expressions and lighting conditions for the same input image, as generated by LEGAN. These input images are randomly selected from the CelebA \cite{celebA} dataset and the results are generated by randomly setting different expression and lighting targets. LEGAN is trained on 33k frontal face images from MultiPIE \cite{MultiPIE} and we do not fine-tune the model on any other dataset. All images are 128$\times$128$\times$3.}
\label{fig:Gal_2}
\end{figure*}

In this section, we share more qualitative results generated by LEGAN on unconstrained data from the AFLW \cite{aflw} and CelebA \cite{celebA} datasets in Figures \ref{fig:Gal_1} and \ref{fig:Gal_2} respectively. The randomly selected input images vary in ethnicity, gender, color composition, resolution, lighting, expression and facial pose. In order to judge LEGAN's generalizability, we only train the model on 33k frontal face images from MultiPIE \cite{MultiPIE} and do not fine tune it on any other dataset.

\section{Recolorization Network: Architecture Details}
For the colorization augmentation network, we use a generator architecture similar to the one used in \cite{SREFI3} for the 128$\times$128$\times$3 resolution. The generator is an encoder-decoder with skip connections connecting the encoder and decoder layers, and the discriminator is the popular CASIANet \cite{CASIA} architecture. Details about the generator layers can be found in Table \ref{Tab:Color_G}.

\begin{table}
\begin{center}
\captionsetup{justification=centering}
\caption{Colorization Generator architecture (input size is 128$\times$128$\times$3)}
\begin{small}
\begin{tabular}{  | c | c| c| }
\hline
\begin{tabular}[x]{@{}c@{}}{\bf Layer}\end{tabular} & \begin{tabular}[x]{@{}c@{}}{\bf Filter/Stride/Dilation}\end{tabular} & \begin{tabular}[x]{@{}c@{}}{\bf \# of filters}\end{tabular}\\
\hline
\hline
  conv0 & 3$\times$3/1/2 & 128 \\
  conv1 & 3$\times$3/2/1 & 64\\
  RB1 & 3$\times$3/1/1 & 64\\
  conv2 & 3$\times$3/2/1 & 128\\
  RB2 & 3$\times$3/1/1 & 128\\
  conv3 & 3$\times$3/2/1 & 256\\
  RB3 & 3$\times$3/1/1 & 256\\
  conv4 & 3$\times$3/2/1 & 512\\
  RB4 & 3$\times$3/1/1 & 512\\
  conv5 & 3$\times$3/2/1 & 1,024 \\
  RB5 & 3$\times$3/1/1 & 1,024\\
  fc1 & 512 & - \\
  fc2 & 16,384 & - \\
  \hline
  \hline
  conv3 & 3$\times$3/1/1 & 4*512\\
  PS1 & - & - \\
  conv4 & 3$\times$3/1/1 & 4*256\\
  PS2 & - & - \\
  conv5 & 3$\times$3/1/1 & 4*128\\
  PS3 & - & - \\
  conv6 & 3$\times$3/1/1 & 4*64\\
  PS4 & - & - \\
  conv7 & 3$\times$3/1/1 & 4*64\\
  PS5 & - & - \\
  conv8 & 5$\times$5/1/1 & 3\\
  \hline
\end{tabular}
\label{Tab:Color_G}
\end{small}
\end{center}
\end{table}

\begin{figure}
\centering
  \includegraphics[width=1.0\linewidth]{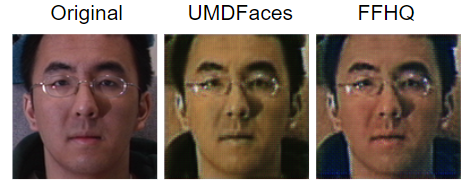}
  \caption{Recolorization Example: We randomly select a test image from the MultiPIE \cite{MultiPIE} dataset and recolor it using the colorization generator snapshots, trained using UMDFaces \cite{UMDFaces} and FFHQ \cite{ProgressiveGAN} datasets respectively. Although the image is recolored, its lighting is preserved by the colorization generator.}
\label{fig:Color_Ex}
\end{figure}

We train two separate versions of the colorization network with randomly selected 10,000 face images from the UMDFaces \cite{UMDFaces} and FFHQ \cite{ProgressiveGAN} datasets. These two trained generators can then be used to augment LEGAN's training set by randomly recoloring the MultiPIE \cite{MultiPIE} images from the training split. Such an example has been shared in Figure \ref{fig:Color_Ex}.

\end{document}